%% file: main_neurips_arxiv.tex
\documentclass{article}

\usepackage[final,nonatbib]{neurips_2020_ml4ps}
\usepackage[utf8]{inputenc} 
\usepackage[T1]{fontenc}    
\usepackage{hyperref}       
\usepackage{url}            
\usepackage{booktabs}       
\usepackage{amsfonts}       
\usepackage{nicefrac}       
\usepackage{microtype}      

\usepackage{amsmath}
\usepackage{algorithm}
\usepackage{algorithmic}
\usepackage{makecell}
\usepackage{color}
\usepackage{graphicx} 
\usepackage{caption}
\usepackage{subcaption}
\usepackage{siunitx}

\renewcommand{\vec}[1]{\mathbf{#1}}

\title{Max-value Entropy Search for Multi-Objective Bayesian Optimization with Constraints}

\author{%
   Syrine Belakaria,
   Aryan Deshwal,
  Janardhan Rao Doppa \\
  School of EECS, Washington State University\\
  \texttt{\{syrine.belakaria, aryan.deshwal, jana.doppa\}@wsu.edu}
}
\begin{document}

\maketitle
\begin{abstract}
  We consider the problem of constrained multi-objective blackbox optimization using expensive function evaluations, where the goal is to approximate the true Pareto set of solutions satisfying a set of constraints while minimizing the number of function evaluations. For example, in aviation power system design applications, we need to find the designs that trade-off total energy and the mass while satisfying specific thresholds for motor temperature and voltage of cells. This optimization requires performing expensive computational simulations to evaluate designs. In this paper, we propose a new approach referred to as {\em Max-value Entropy Search for Multi-objective Optimization with Constraints (MESMOC)} to solve this problem. MESMOC employs an output-space entropy based acquisition function to efficiently select the sequence of inputs for evaluation to uncover high-quality pareto-set solutions while satisfying constraints.
 We apply MESMOC to two real-world engineering design applications to demonstrate its effectiveness.
\end{abstract}

\input{Introduction.tex}

\input{Problem_setup.tex}

\input{OurApproach.tex}

\input{Experiments.tex}

\vspace{1.0ex}

\noindent {\bf Acknowledgements.} The authors gratefully acknowledge the support from National Science Foundation (NSF) grants IIS-1845922 and OAC-1910213. The views expressed are those of the authors and do not reflect the official policy or position of the NSF.


\input{main_neurips_arxiv.bbl}

\end{document}

%% file: Introduction.tex

\section{Introduction}


Many engineering and scientific applications involve making design choices to optimize multiple objectives. Some examples include tuning the knobs of a compiler to optimize performance and efficiency of a set of software programs;  and designing new materials to optimize strength, elasticity, and durability. There are three common challenges in solving this kind of multi-objective optimization (MO) problems: {\bf 1)} The objective functions are unknown and we need to perform expensive experiments to evaluate each candidate design choice. For example, performing computational simulations and physical lab experiments for compiler optimization and material design applications respectively. {\bf 2)} The objectives are conflicting in nature and all of them cannot be optimized simultaneously. {\bf 3)} The problem involves several black-box constraints that need to be satisfied. Therefore, we need to find the {\em Pareto optimal} set of solutions satisfying the constraints. A solution is called Pareto optimal if it cannot be improved in any of the objectives without compromising some other objective. The overall 
goal is to approximate the true Pareto set satisfying the constraints while minimizing the number of function evaluations. 

Bayesian Optimization (BO) (\cite{shahriari2016taking}) is an effective framework to solve blackbox optimization problems with expensive function evaluations. The key idea behind BO is to
build a cheap surrogate model (e.g., Gaussian Process (\cite{williams2006gaussian}) using the real experimental evaluations; and employ it to intelligently select the sequence of function evaluations using an acquisition function, e.g., expected improvement (EI). There is a large body of literature on single-objective BO algorithms (\cite{shahriari2016taking,BOCS,PSR,DBO}) and their applications including hyper-parameter tuning of machine learning methods (\cite{snoek2012practical,kotthoff2017auto}). However, there is relatively less work on the more challenging problem of BO for multiple objectives (\cite{knowles2006parego,emmerich2008computation,PESMO,MESMO,USEMO}) and very limited prior work to address constrained MO problems (\cite{garrido2019predictive,feliot2017bayesian}). PESMOC (\cite{garrido2019predictive}) is the current state-of-the-art method in this problem setting. PESMOC is an information-theoretic approach that relies on the principle of input space entropy search. However, it is computationally expensive to optimize the acquisition function behind PESMOC. A series of approximations are performed to improve the efficiency potentially at the expense of accuracy.

In this paper, we propose a new and principled approach referred to as {\em {\bf M}ax-value {\bf E}ntropy {\bf S}earch for {\bf M}ulti-objective {\bf O}ptimization with {\bf C}onstraints} (MESMOC). MESMOC employs an {\em output space entropy} based acquisition function to select the candidate inputs for evaluation. The key idea is to evaluate the input that maximizes the information gain about the optimal Pareto front in each iteration while satisfying the constraints. Output space entropy search has many advantages over algorithms based on input space entropy search (\cite{MESMO}): a) allows tighter approximation ; b) significantly cheaper to compute; and c) naturally lends itself to robust optimization. MESMOC is an extension of the MESMO algorithm \cite{MESMO} based on output space information gain, which was shown to be  efficient and robust, to the challenging constrained MO setting.

%% file: Problem_setup.tex
\section{Background and Problem Setup}

\noindent {\bf Bayesian Optimization (BO) Framework.} BO is a very efficient framework to solve global optimization problems using {\em black-box evaluations of expensive objective functions}. Let $\mathfrak{X} \subseteq \Re^d$ be an input space. In single-objective BO formulation, we are given an unknown real-valued objective function $f: \mathfrak{X} \mapsto \Re$, which can evaluate each input $\vec{x} \in \mathfrak{X}$ to produce an evaluation $y$ = $f(\vec{x})$.  Each evaluation $f(\vec{x})$ is expensive in terms of the consumed resources. The main goal is to find an input $\vec{x^*} \in \mathfrak{X}$ that approximately optimizes $f$ by performing a limited number of function evaluations. BO algorithms learn a cheap surrogate model from training data obtained from past function evaluations. They intelligently select the next input for evaluation by trading-off exploration and exploitation to quickly direct the search towards optimal inputs. The three key elements of BO framework are:

\hspace{2.0ex} {\bf 1) Statistical Model} of the true function $f(x)$. {\em Gaussian Process (GP)} \cite{williams2006gaussian} is the most commonly used model. A GP over a space $\mathfrak{X}$ is a random process from $\mathfrak{X}$ to $\Re$. It is characterized by a mean function $\mu : \mathfrak{X} \mapsto \Re$ and a covariance or kernel function $\kappa : \mathfrak{X} \times \mathfrak{X} \mapsto \Re$. If a function $f$ is sampled from GP($\mu$, $\kappa$), then $f(x)$ is distributed normally $\mathcal{N}(\mu(x), \kappa(x,x))$ for a finite set of inputs from $x \in \mathcal{X}$.

\hspace{2.0ex} {\bf 2) Acquisition Function} ($\alpha$) to score the utility of evaluating a candidate input $\vec{x} \in \mathfrak{X}$ based on the statistical model. Some popular acquisition functions in the single-objective literature include expected improvement (EI), upper confidence bound (UCB), predictive entropy search (PES) \cite{PES}, and max-value entropy search (MES) \cite{MES}.

\hspace{2.0ex} {\bf 3) Optimization Procedure} to select the best scoring candidate input according to $\alpha$ depending on statistical model. DIRECT \cite{jones1993lipschitzian} is a very popular approach for acquisition function optimization.

\vspace{1.5ex}

\noindent {\bf Multi-Objective Optimization Problem with Constraints.}  Without loss of generality, our goal is to minimize real-valued objective functions $f_1(\vec{x}), f_2(\vec{x}),\cdots,f_K(\vec{x})$, with  $K \geq 2$,  while satisfying $L$ black-box constraints of the form $C_1(x)\geq 0, C_2(x)\geq 0,\cdots,C_L(x)\geq 0$ over continuous space  $\mathfrak{X} \subseteq \Re^d$. Each evaluation of an input $\vec{x}\in \mathfrak{X}$ produces a vector of objective values and constraint values $\vec{y}$ = $(y_{f_1}, y_{f_2},\cdots,y_{f_K},y_{c_1} \cdots y_{c_L})$ where $y_{f_j} = f_j(x)$ for all $j \in \{1,2, \cdots, K\}$ and $y_{c_i} = C_i(x)$ for all $i \in \{1,2, \cdots, L\}$. We say that a valid point $\vec{x}$ (satisfies all constraints) {\em Pareto-dominates} another point $\vec{x'}$ if $f_j(\vec{x}) \leq f_j(\vec{x'}) \hspace{1mm} \forall{j}$ and there exists some $j \in \{1, 2, \cdots,K\}$ such that $f_j(\vec{x}) < f_j(\vec{x'})$. The optimal solution of MOO problem with constraints is a set of points $\mathcal{X}^* \subset \mathfrak{X}$ such that no point $\vec{x'} \in \mathfrak{X} \setminus \mathcal{X}^*$ Pareto-dominates a point $\vec{x} \in \mathcal{X}^*$ and all points in $\mathcal{X}^*$ satisfy the problem constraints. The solution set $\mathcal{X}^*$ is called the optimal {\em Pareto set} and the corresponding set of function values $\mathcal{Y}^*$  is called the optimal {\em Pareto front}. Our goal is to approximate $\mathcal{X}^*$ by minimizing the number of function evaluations.

%% file: OurApproach.tex

\section{MESMOC for Multi-Objective Optimization with Constraints}
\label{section4}


In this section, we explain the technical details of our proposed MESMOC algorithm. We first mathematically describe the output space entropy based acquisition function and provide an algorithmic approach to efficiently compute it.

\vspace{1.0ex}

\noindent {\bf Surrogate models.} Gaussian processes (GPs) are shown to be effective surrogate models in prior work on single and multi-objective BO \cite{PES,EST,MES,gp-ucb,PESMO}. Similar to prior work \cite{PESMO}, we model the objective functions and blackbox constraints by independent GP models $\mathcal{M}_{f_1},\mathcal{M}_{f_2},\cdots,\mathcal{M}_{f_K}$ and $\mathcal{M}_{c_1},\mathcal{M}_{c_2},\cdots,\mathcal{M}_{f_K}$ with zero mean and i.i.d. observation noise. Let $\mathcal{D} = \{(\vec{x}_i, \vec{y}_i)\}_{i=1}^{t-1}$ be the training data from past $t{-1}$ function evaluations, where  
$\vec{x}_i \in \mathfrak{X}$ is an input and $\vec{y}_i = \{y^i_{f_1},\cdots,y^i_{f_K},y^i_{c_1},\cdots y^i_{c_L}\}$ is the output vector resulting from evaluating the objective functions and constraints at $\vec{x}_i$. We learn surrogate models from $\mathcal{D}$.

\vspace{1.0ex}

\noindent{\bf Output space entropy based acquisition function.} Input space entropy based methods such as PESMO \cite{PESMO} selects the next candidate input $\vec{x}_{t}$ (for ease of notation, we drop the subscript in below discussion) by maximizing the information gain about the optimal Pareto set $\mathcal{X}^*$. The acquisition function based on input space entropy is given as follows:
\begin{align}
    \alpha(\vec{x}) &= I(\{\vec{x}, \vec{y}\}, \mathcal{X}^* \mid D) \label{eqn_orig_inf_gain}\\ 
    &= H(\mathcal{X}^* \mid D) - \mathbb{E}_y [H(\mathcal{X}^* \mid D \cup \{\vec{x}, \vec{y}\})] \label{eqn_exp_redn} \\
    &= H(\vec{y} \mid D, \vec{x}) - \mathbb{E}_{\mathcal{X}^*} [H(\vec{y} \mid D,   \vec{x}, \mathcal{X}^*)] \label{eqn_symmetric_pesmo}
\end{align} 

Information gain is defined as the expected reduction in entropy $H(.)$ of the posterior distribution $P(\mathcal{X}^* \mid D)$ over the optimal Pareto set $\mathcal{X}^*$ as given in Equations \ref{eqn_exp_redn} and \ref{eqn_symmetric_pesmo} (resulting from symmetric property of information gain). This mathematical formulation relies on a very expensive and high-dimensional ($m\cdot d$ dimensions) distribution  $P(\mathcal{X}^* \mid D)$, where $m$ is size of the optimal Pareto set $\mathcal{X}^*$. Furthermore, optimizing the second term in r.h.s poses significant challenges: a) requires a series of approximations \cite{PESMO} which can be potentially sub-optimal; and b) optimization, even after approximations, is expensive c) performance is strongly dependent on the number of Monte-Carlo samples.

To overcome the above challenges of computing input space entropy based acquisition function,  \cite{MESMO} proposed to maximize the information gain about the optimal {\bf Pareto front}  $\mathcal{Y}^*$. However, MESMO did not address the challenge of constrained Pareto front. We propose an extension of MESMO's acquisition function to maximize the information gain between the next candidate input for evaluation $\vec{x}$ and constrained Pareto front $\mathcal{Y}^*$ given as:
\begin{align}
        \alpha(\vec{x}) &= I(\{\vec{x}, \vec{y}\}, \mathcal{Y}^* \mid D) \\ 
    &= H(\mathcal{Y}^* \mid D) - \mathbb{E}_y [H(\mathcal{Y}^* \mid D \cup \{\vec{x}, \vec{y}\})]  \\
    &= H(\vec{y} \mid D, \vec{x}) - \mathbb{E}_{\mathcal{Y}^*} [H(\vec{y} \mid D,   \vec{x}, \mathcal{Y}^*)] \label{eqn_symmetric_mesmo}
\end{align}
In this case, the output vector $\vec{y}$ is $K+L$ dimensional: $\vec{y}$ = $(y_{f_1}, y_{f_2},\cdots,y_{f_K},y_{c_1} \cdots y_{c_L})$ where $y_{f_j}$ = $f_j(x)$ for all $j \in \{1,2, \cdots, K\}$ and $y_{c_i}$ = $C_i(x)$ for all $i \in \{1,2, \cdots, L\}$. 
Consequently, the first term in the  r.h.s of equation \ref{eqn_symmetric_mesmo}, entropy of a factorizable $(K+L)$-dimensional Gaussian distribution $P(\vec{y}\mid D, \vec{x}$, can be computed in closed form as shown below:
\begin{align}
    H(\vec{y} \mid D, \vec{x}) = \frac{(K+C)(1+\ln(2\pi))}{2} +  \sum_{j = 1}^K  \ln (\sigma_{f_j}(\vec{x}))+  \sum_{i = 1}^L  \ln (\sigma_{c_i}(\vec{x})) \label{eqn_unconditioned_entropy}
\end{align}
where $\sigma_{f_j}^2(\vec{x})$ and  $\sigma_{c_i}^2(\vec{x})$ are the predictive variances of $j^{th}$ function and $i^{th}$ constraint GPs respectively at input $\vec{x}$. The second term in the r.h.s of equation \ref{eqn_symmetric_mesmo} is an expectation over the Pareto front $\mathcal{Y}^*$. We can approximately compute this term via Monte-Carlo sampling as shown below: 

\begin{align}
    \mathbb{E}_{\mathcal{Y}^*} [H(\vec{y} \mid D,   \vec{x}, \mathcal{Y}^*)] \simeq \frac{1}{S} \sum_{s = 1}^S [H(\vec{y} \mid D,   \vec{x}, \mathcal{Y}^*_s)] \label{eqn_summation}
\end{align}
where $S$ is the number of samples and $\mathcal{Y}^*_s$ denote a sample Pareto front. The main advantages of our acquisition function are: computational efficiency and robustness to the number of samples (\cite{MESMO}). 

There are two key algorithmic steps to compute Equation \ref{eqn_summation}: 1) How to compute Pareto front samples $\mathcal{Y}^*_s$?; and 2) How to compute the entropy with respect to a given Pareto front sample $\mathcal{Y}^*_s$? We provide solutions for these two questions below.

\vspace{1.0ex}

\hspace{2.0ex} {\bf 1) Computing Pareto front samples via cheap multi-objective optimization.} To compute a Pareto front sample $\mathcal{Y}^*_s$, we first sample functions and constraints from the posterior GP models via random fourier features (\cite{PES,random_fourier_features}) and then solve a cheap multi-objective optimization over the $K$ sampled functions and $L$ sampled constraints.

\hspace{3.5ex}{\em Cheap MO solver.} We sample $\Tilde{f}_i$ from GP model $\mathcal{M}_{f_j}$ for each of the $K$ functions and $\Tilde{C}_i$ from GP model $\mathcal{M}_{c_i}$ for each of the $L$ constraints. A {\em cheap} constrained multi-objective optimization problem over the $K$ sampled functions  $\Tilde{f}_1,\Tilde{f}_2,\cdots,\Tilde{f}_k$ and the $L$ sampled constraints  $\Tilde{C}_1,\Tilde{C}_2,\cdots,\Tilde{C}_L$ is solved to compute the sample Pareto front $\mathcal{Y}^*_s$. This cheap multi-objective optimization also allows us to capture the interactions between different objectives while satisfying the constraints. We employ the popular constrained NSGA-II algorithm (\cite{deb2002nsga,deb2002fast}) to solve the constrained MO problem with cheap objective functions noting that any other algorithm can be used to similar effect.

\vspace{1.0ex}

\hspace{2.0ex}{\bf 2) Entropy computation with a sample Pareto front.}
Let $\mathcal{Y}^*_s = \{\vec{z}^1, \cdots, \vec{z}^m \}$ be the sample Pareto front,  where $m$ is the size of the Pareto front and each $\vec{z}^i$ is a $(K+L)$-vector evaluated at the $K$ sampled functions and $L$ sampled constraints $\vec{z}^i = \{z^i_{f_1},\cdots,z^i_{f_K},z^i_{c_1},\cdots,z^i_{c_L}\}$. The following inequality holds for each component $y_j$ of the $(K+L)$-vector $\vec{y} = \{y_{f_1},\cdots,y_{f_K},y_{c_1},\cdots y_{c_L}\}$ in the entropy term $H(\vec{y} \mid D,   \vec{x}, \mathcal{Y}^*_s)$:
\begin{align}
 y_j &\leq \max \{z^1_j, \cdots z^m_j \} \quad \forall j \in \{f_1,\cdots,f_K,c_1,\cdots,c_L\} \label{inequality}
\end{align}

The inequality essentially says that the $j^{th}$ component of $\vec{y}$ (i.e., $y_j$) is upper-bounded by a value obtained by taking the maximum of $j^{th}$ components of all $m$ $(K+L)$-vectors in the Pareto front $\mathcal{Y}^*_s$. This inequality had been proven by a contradiction in (\cite{MESMO}) for $j \in \{f_1,\cdots,f_K\}$. We assume the same  for $j \in \{c_1,\cdots,c_L\}$.

By combining the inequality \ref{inequality} and the fact that each function is modeled as an independent GP, we can model each component $y_j$ as a truncated Gaussian distribution since the distribution of $y_j$ needs to satisfy $ y_j \leq \max \{z^1_j, \cdots z^m_j \}$.  Furthermore, a common property of entropy measure allows us to decompose the entropy of a set of independent variables into a sum over entropies of individual variables \cite{information_theory}:
\begin{align}
H(\vec{y} \mid D,   \vec{x}, \mathcal{Y}^*_s) \simeq \sum_{j=1}^K H(y_{f_j}|D, \vec{x}, \max \{z^1_{f_j}, \cdots z^m_{f_j} \}) +\sum_{i=1}^C H(y_{c_i}|D, \vec{x}, \max \{z^1_{c_i}, \cdots z^m_{c_i} \})  \label{eqn_sep_ineq}
\end{align}
The r.h.s is a summation over entropies of  $(K+L)$-variables $\vec{y} = \{y_{f_1},\cdots,y_{f_K},y_{c_1},\cdots y_{c_L}\}$.
The differential entropy for each $y_j$ is the entropy of a truncated Gaussian distribution (\cite{entropy_handbook}) and given by the following equations:
\begin{align}
    H(y_{f_j}|D, \vec{x}, y_s^{f_j*}) &\simeq  \left[\frac{(1 + \ln(2\pi))}{2}+  \ln(\sigma_{f_j}(\vec{x})) +  \ln \Phi(\gamma_s^{f_j}(\vec{x})) - \frac{\gamma_s^{f_j}(\vec{x}) \phi(\gamma_s^{f_j}(\vec{x}))}{2\Phi(\gamma_s^{f_j}(\vec{x}))}\right] \label{eqn_entropy_fj}
  \end{align}
\begin{align}
   H(y_{c_i}|D, \vec{x}, y_s^{c_i*})&\simeq  \left[\frac{(1 + \ln(2\pi))}{2}+  \ln(\sigma_{c_i}(\vec{x})) +  \ln \Phi(\gamma_s^{c_i}(\vec{x})) - \frac{\gamma_s^{c_i}(\vec{x}) \phi(\gamma_s^{c_i}(\vec{x}))}{2\Phi(\gamma_s^{c_i}(\vec{x}))}\right]    
    \label{eqn_entropy_cj}
\end{align} 
Consequently we have: 
\begin{align}
    H(\vec{y} \mid D,   \vec{x}, \mathcal{Y}^*_s) &\simeq \sum_{j=1}^K \left[\frac{(1 + \ln(2\pi))}{2}+  \ln(\sigma_{f_j}(\vec{x})) +  \ln \Phi(\gamma_s^{f_j}(\vec{x})) - \frac{\gamma_s^{f_j}(\vec{x}) \phi(\gamma_s^{f_j}(\vec{x}))}{2\Phi(\gamma_s^{f_j}(\vec{x}))}\right] \nonumber\\
&+ \sum_{i=1}^L \left[\frac{(1 + \ln(2\pi))}{2}+  \ln(\sigma_{c_i}(\vec{x})) +  \ln \Phi(\gamma_s^{c_i}(\vec{x})) - \frac{\gamma_s^{c_i}(\vec{x}) \phi(\gamma_s^{c_i}(\vec{x}))}{2\Phi(\gamma_s^{c_i}(\vec{x}))}\right]    
    \label{eqn_entropy_closed}
\end{align}

where $\gamma_s^{c_i}(x) = \frac{y_s^{c_i*} - \mu_{c_i}(\vec{x})}{\sigma_{c_i}(\vec{x})}$, $\gamma_s^{f_j}(x) = \frac{y_s^{f_j*} - \mu_{f_j}(\vec{x})}{\sigma_{f_j}(\vec{x})}$, $y_s^{c_i*}$ and $y_s^{f_j*}$ are the maximum values of constraint $\Tilde{c_i}$  and function $\Tilde{f_j}$ reached after the cheap multi-objective optimization over sampled functions and constraints:
$y_s^{c_i*} = \max \{z^1_{c_i}, \cdots z^m_{c_i} \}$, $y_s^{f_j*} = \max \{z^1_{f_j}, \cdots z^m_{f_j} \}$, 
$\phi$ and $\Phi$ are the p.d.f and c.d.f of a standard normal distribution respectively. By combining equations \ref{eqn_unconditioned_entropy} and \ref{eqn_entropy_closed} with Equation \ref{eqn_symmetric_mesmo}, we get the final form of our acquisition function as shown below:
\begin{align}
\alpha(\vec{x}) \simeq \frac{1}{S} \sum_{s=1}^S\left[ \sum_{j=1}^K  \frac{\gamma_s^{f_j}(\vec{x}) \phi(\gamma_s^{f_j}(\vec{x}))}{2\Phi(\gamma_s^{f_j}(\vec{x}))} - \ln \Phi(\gamma_s^{f_j}(\vec{x}))  + \sum_{i=1}^L  \frac{\gamma_s^{c_i}(\vec{x}) \phi(\gamma_s^{c_i}(\vec{x}))}{2\Phi(\gamma_s^{c_i}(\vec{x}))} - \ln \Phi(\gamma_s^{c_i}(\vec{x})) \right] \label{eqn_final}
 \end{align} 
A complete description of the MESMOC algorithm is given in Algorithm \ref{alg:MESMO}. The blue colored steps correspond to computation of our output space entropy based acquisition function via sampling.

\begin{algorithm}[h]
\caption{MESMOC Algorithm}
\label{alg:MESMO}
\textbf{Input}: input space $\mathfrak{X}$; $K$ blackbox functions $f_1(x), f_2(x),\cdots,f_K(x)$; $L$ blackbox constraints $C_1(x), C_2(x),\cdots,C_L(x)$; and  maximum no. of iterations $T_{max}$
\begin{algorithmic}[1] 
\STATE Initialize Gaussian process models $\mathcal{M}_{f_1}, \mathcal{M}_{f_2},\cdots, \mathcal{M}_{f_K}$ and $\mathcal{M}_{c_1}, \mathcal{M}_{c_2},\cdots, \mathcal{M}_{c_L}$ by evaluating at $N_0$ initial points
\FOR{each iteration $t$ = $N_0+1$ to $T_{max}$}
\STATE Select $\vec{x}_{t} \leftarrow \arg max_{\vec{x}\in \mathfrak{X}} \hspace{2 mm} \alpha_t(\vec{x}) $
 \\ \qquad \qquad \textbf{s.t} $(\mu_{c_1}\geq 0, \cdots,\mu_{c_L}\geq 0 )$
 \STATE $\alpha_t(.)$ is computed as:
 \STATE \quad \textcolor{blue}{for each sample $s \in {1,\cdots,S}$:} 
 \STATE \quad \quad \textcolor{blue}{Sample $\Tilde{f}_j \sim \mathcal{M}_{f_j}, \quad \forall{j \in \{1,\cdots, K\}} $}
  \STATE \quad \quad \textcolor{blue}{Sample $\Tilde{C}_i \sim \mathcal{M}_{c_i}, \quad \forall{i \in \{1,\cdots, L\}} $}
 \STATE \quad \quad \textcolor{blue}{// Solve {\em cheap} MOO over $(\Tilde{f}_1, \cdots, \Tilde{f}_K)$ constrained by $(\Tilde{C}_1, \cdots, \Tilde{C}_L)$}
 \STATE \quad \quad \textcolor{blue}{$\mathcal{Y}_s^* \leftarrow \arg max_{x \in \mathcal{X}} (\Tilde{f}_1, \cdots, \Tilde{f}_K)$ \\ \qquad \qquad \textbf{s.t} $(\Tilde{C}_1\geq 0, \cdots, \Tilde{C}_L\geq 0)$} 
 \STATE  \quad \textcolor{blue}{Compute $\alpha_t$(.) based on the $S$ samples of $\mathcal{Y}_s^*$ as given in Equation \ref{eqn_final}}
\STATE Evaluate $\vec{x}_{t}$; $\vec{y}_{t} \leftarrow (f_1(\vec{x}_{t}),\cdots,f_K(\vec{x}_{t}),C_1(\vec{x}_{t}),\cdots,C_L(\vec{x}_{t}))$ 
\STATE Aggregate data: $\mathcal{D} \leftarrow \mathcal{D} \cup \{(\vec{x}_{t}, \vec{y}_{t})\}$ 
\STATE Update models $\mathcal{M}_{f_1}, \mathcal{M}_{f_2},\cdots, \mathcal{M}_{f_K}$ and $\mathcal{M}_{c_1}, \mathcal{M}_{c_2},\cdots, \mathcal{M}_{c_L}$
\STATE $t \leftarrow t+1$
\ENDFOR
\STATE \textbf{return} Pareto front of $f_1(x), f_2(x),\cdots,f_K(x)$ based on $\mathcal{D}$
\end{algorithmic}
\end{algorithm}

%% file: Experiments.tex

 \section{Experiments and Results}
 
 
In this section, we describe our experimental evaluation of MESMOC on two real-world engineering applications, namely, electrified aviation power system design and analog circuit design optimization tasks.

\subsection{Experimental Setup}

We compare MESMOC with PESMOC (\cite{garrido2019predictive}). Due to lack of BO approaches for constrained MO, we compare to known genetic algorithms (NSGA-II and MOEAD). However, they require a large number of function evaluations to converge which is not practical for optimization of expensive functions. We use a GP based statistical model with squared exponential (SE) kernel in all our experiments. 
The hyper-parameters are estimated after every 5 function evaluations. We initialize the GP models for all functions by sampling initial points at random.
The code is available at (github.com/belakaria/MESMOC)

\vspace{1.0ex}

{\bf Electrified aviation power system design.}  We consider optimizing the design of electrified aviation power system of unmanned aerial vehicle (UAV) via a time-based static simulation. The UAV system architecture consists of a central Li-ion
battery pack, hex-bridge DC-AC inverters, PMSM motors, and necessary wiring (\cite{belakaria2020machine}). Each candidate input consists of a set of $5$ ($d$=5) variable design parameters such as the battery pack configuration (battery cells in series, battery cells in parallel) and motor
size (number of motors, motor stator winding length, motor stator winding turns). We minimize two objective functions: mass and total energy. This problem has $5$ black-box constraints: 
\begin{align}
    &{C}_0: \text{Maximum final depth of discharge} \leq 75 \% \nonumber \\
    &{C}_1: \text{Minimum cell voltage}\geq 3 V \nonumber\\
    &{C}_2: \text{Maximum motor temperature} \leq \ang{125} C \nonumber\\
    &{C}_3:\text{Maximum inverter temperature} \leq \ang{120} C \nonumber\\
    & {C}_5:\text{ Maximum modulation index} \leq 1.3 \nonumber
\end{align}
For a design to be valid, the simulated UAV must be capable of completing the
specified mission without violating any of the constraints. The overall design space has a total of 250,000 possible candidate designs. Out of the entire design space, only 9\% of the designs are valid (i.e., satisfy all the constraints), which makes it a very challenging task. Additionally, only five points are in the optimal Pareto front. 

\vspace{1.0ex}

\noindent {\bf Analog circuit optimization domain.} We consider optimizing the design of a multi-output switched-capacitor voltage regulator via Cadence circuit simulator that imitates the real hardware \cite{DATE-2020}. This circuit relies on a dynamic frequency switching clock. Each candidate circuit design is defined by 33 input variables ($d$=33). The first 24 variables are the width, length, and unit of the eight capacitors of the circuit $W_i,L_i,M_i ~ \forall i \in 1\cdots 8$. The remaining input variables are four output voltage references $V_{ref_i}~ \forall i \in 1\cdots 4$ and four resistances $R_i ~ \forall i \in 1\cdots 4$ and a switching frequency $f$. We optimize nine objectives: maximize efficiency $Eff$, maximize four output voltages $V_{o_1} \cdots V_{o_4}$, and minimize four output ripples $OR_{1} \cdots OR_{4}$.
Our problem has a total of nine constraints. Since some of the constraints have upper bounds and lower bounds, they are defined in the problem by 15 different constraints:
\begin{align}
    &{C}_0: Cp_{total} \simeq 20 nF  ~ with ~ Cp_{total}= \sum_{i=1}^8 (1.955W_iL_i+0.54(W_i+L_i))M_i \nonumber \\
    &{C}_1~ to~ {C}_4:  V_{o_i}\geq V_{ref_i} ~ \forall \in{1\cdots4} \nonumber \\
    &{C}_5\ to\ {C}_{8\ }:\ \ OR_{lb}\le OR_i\le OR_{ub} ~ \forall i\ \in{1\cdots4} \nonumber\\
    &{C}_9:Eff \le 100\% \nonumber
\end{align}
where $OR_{lb}$ and $OR_{ub}$ are the predefined lower-bound and upper-bound of $OR_i$ respectively. $Cp_{total}$ is the total capacitance of the circuit.

\subsection{Results and Discussion}
\begin{figure}[h!] 
    \centering

    \begin{minipage}{0.49\textwidth}
        \centering
        \includegraphics[width=0.89\textwidth]{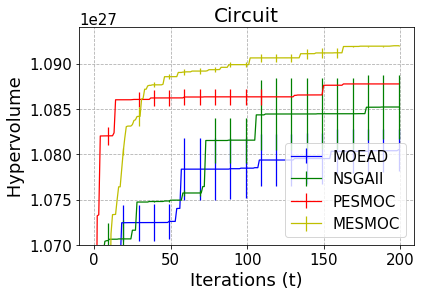} 
    \end{minipage}\hfill
    \begin{minipage}{0.49\textwidth}
        \centering
        \includegraphics[width=0.86\textwidth]{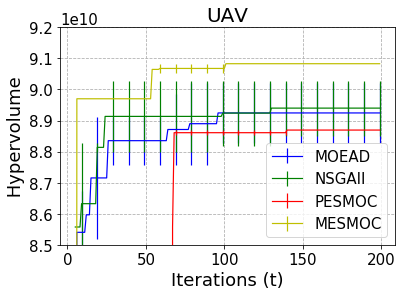} 
    \end{minipage}
\caption{Results of different constrained multi-objective algorithms including MESMOC. The hypervolume metric is shown as a function of the number function evaluations.}\label{fig:hv}
    \end{figure}

We evaluate the performance of our algorithm and the baselines using the Pareto hypervolume (PHV) metric. PHV is a commonly employed metric to measure the quality of a given Pareto front \cite{zitzler1999evolutionary}. 
Figure \ref{fig:hv} shows that MESMOC outperforms existing baselines. It recovers a better Pareto front with a significant gain in the number of function evaluations.
Both of these experiments are motivated by real-world engineering applications where further analysis of the designs in the Pareto front is crucial. 

\vspace{1.0ex}

\noindent {\bf Electrified aviation power system design.}
In this setting, the input space is discrete with 250,000 combinations of design parameters. Out of the entire design space, only 9\% of design combinations passed all the constraints and only five points are in the optimal Pareto front. From a domain expert perspective, satisfying all the constraints is critical. Hence, the results reported for the hypervolume include only points that satisfy all the constraints. Despite the hardness of the problem, 50\% of the designs selected by MESMOC satisfy all the constraints while for PESMOC, MOEAD, and NSGA-II, this was 1.5\%,  9.5\%, and 7.5\% respectively. MESMOC was not able to recover all the five points of the optimal Pareto front. However, it was able to reach closely approximate the true Pareto front and recover better designs than the baselines.

\vspace{1.0ex}

\noindent {\bf Analog circuit design optimization.} In this setting, the input space is continuous, consequently there is an infinite number of candidate designs. From a domain expert perspective, satisfying all the constraints is not critical and is impossible to achieve. The main goal is to satisfy most of the constraints (and getting close to satisfying the threshold for violated constraints) while reaching the best possible objective values. Therefore, the results reported for the hypervolume include all the evaluated points. In this experiment, the efficiency of circuit is the most important objective function. The table in Figure 2 shows the optimized circuit parameters from different algorithms.  

    \begin{figure}[h!]
  \centering
  \includegraphics[width=.6\linewidth]{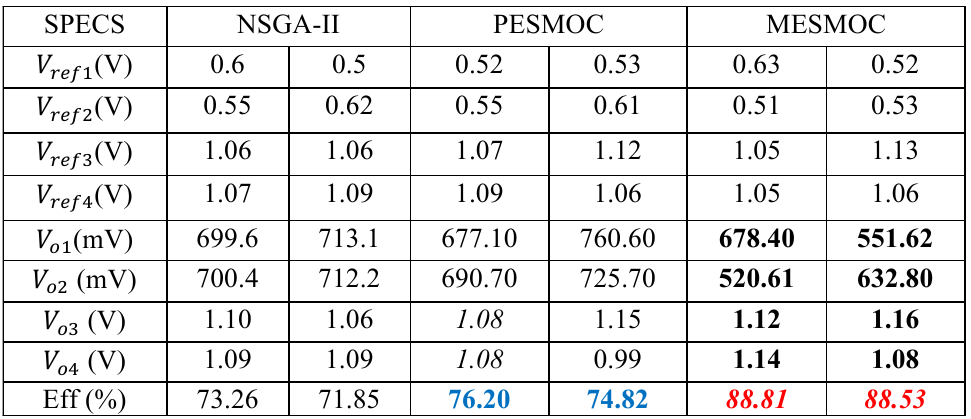}

\caption{Comparison table of optimized circuit parameters obtained with different algorithms (designs are selected from the Pareto set prioritized by efficiency)}  \label{fig:table}
\end{figure}

All algorithms can generate design parameters for the circuit that meets the voltage reference requirements. The optimized circuit using MESMOC can achieve the highest conversion efficiency of 88.81\% (12.61\% improvement when compared with PESMOC and 17.86\% improvement when compared with NSGA-II) with similar output ripples.  The circuit with optimized parameters can generate the target output voltages within the range of 0.52V to 0.76V (1/3x ratio) and 0.99V to 1.17V (2/3x ratio) under the loads varying from 14 Ohms to 1697 Ohms. 